\pdfoutput=1

\documentclass[11pt]{article}

\usepackage[final]{acl}

\usepackage{times}
\usepackage{latexsym}

\usepackage[T1]{fontenc}

\usepackage[utf8]{inputenc}

\usepackage{microtype}

\usepackage{inconsolata}

\usepackage{hyperref}       
\usepackage{url}            
\usepackage{booktabs}       
\usepackage{amsfonts}       
\usepackage{nicefrac}       

\usepackage{xcolor}         
\usepackage{natbib}

\usepackage{amsthm,amsmath,amssymb}
\usepackage{mathrsfs}
\usepackage{enumerate}
\usepackage{float}
\usepackage{bbding}
\usepackage{nicematrix}
\usepackage{arydshln}
\usepackage{balance}
\usepackage{multirow}
\usepackage{subfig}

\newtheorem{definition}{Definition}

\title{Robust Knowledge Graph Embedding via Denoising}

\author{
  Tengwei Song \quad Xudong Ma \quad Yang Liu \quad Jie Luo \\
  Beihang University, Beijing, China \\
  \texttt{\{songtengwei, luojie\}@buaa.edu.cn}
}

\begin{document}
\maketitle
\begin{abstract}
We focus on obtaining robust knowledge graph embedding under perturbation in the embedding space. To address these challenges, we introduce a novel framework, Robust Knowledge Graph Embedding via Denoising, which enhances the robustness of KGE models on noisy triples. By treating KGE methods as energy-based models, we leverage the established connection between denoising and score matching, enabling the training of a robust denoising KGE model. Furthermore, we propose certified robustness evaluation metrics for KGE methods based on the concept of randomized smoothing. Through comprehensive experiments on benchmark datasets, our framework consistently shows superior performance compared to existing state-of-the-art KGE methods when faced with perturbed entity embedding. 
\end{abstract}

\section{Introduction}
Despite the success of knowledge graph embedding (KGE) models in capturing complex relation patterns in Knowledge Graphs (KGs), they remain vulnerable to noisy or incomplete triples, which can lead to inaccurate predictions \citep{kgenoise2018}. Enhancing the robustness of KGE models is crucial, especially in applications like semantic search and recommendation systems, where reliability and accurate reasoning are essential \citep{madry2019deeplearningmodelsresistant}.

Robustness of knowledge graphs(KGs) in existing works focuses on dealing with noise in data space, where noise in KGs is manifested as incorrect triples, missing relations, or spurious connections  \citep{kgenoise2018,kgenoise2023}. 
In recent years, inspired by the growing interest in the area of embedding space perturbations for enhancing robustness in NLP models \citep{lee2021learning,nlp2023robustness,asl-etal-2023-robustembed}, we aim to explore the robustness of KGE methods under perturbed embedding. 

\begin{figure}
    \centering
   \subfloat[Illustration]{\includegraphics[height=3.5cm]{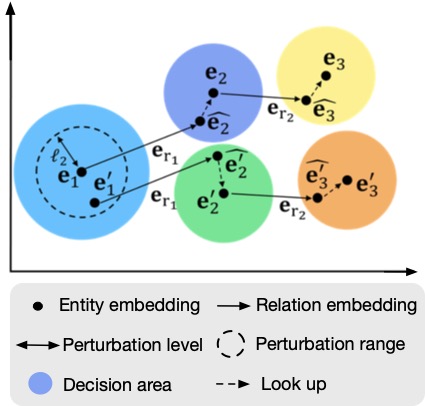}\label{fig:robust_kge}}
   \subfloat[Empirical result]{\includegraphics[height=3.5cm]       {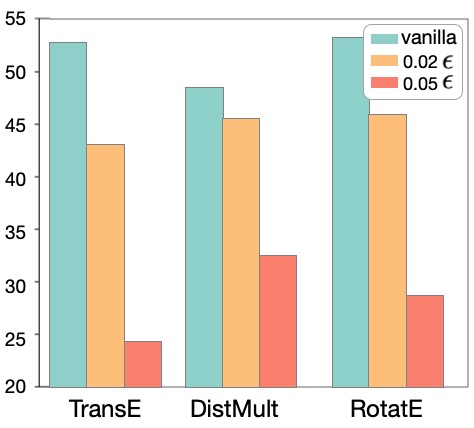}
   \label{fig:intro_noise_bar}}
    \caption{Link prediction shift caused by embedding level perturbation.}
    \label{fig:kge_robustness_intro}
\end{figure}

When the robustness of a KGE method is limited, perturbing an entity can shift link prediction results. As Figure \ref{fig:robust_kge} shows, applying \( l_2 \) perturbations to entity \( e_1 \) can cause its link prediction results to deviate from the original correct entity \( e_2 \) and fall into the decision region of entity \( e'_2 \). This error propagates through link prediction inference as the number of hops increases, severely affecting the results of downstream tasks such as multi-hop reasoning.
We conduct an empirical evaluation on link prediction by adding two scales of Gaussian noise to the embedding of entities.
Figure \ref{fig:intro_noise_bar} shows that the performance on the Hit@10 (\%) ratio in link prediction declines severely when adding minor noise.

In this paper, we extend our inquiry into the resilience of KGE methods against embedding distortions. Specifically, we introduce a novel denoising framework designed to reinforce the robustness of KGE models under embedding perturbations. Our approach utilizes the principles of energy-based models and score matching, which are instrumental in training KGE models that can effectively denoise and recover from such perturbations. Additionally, we propose a new set of certified robustness evaluation metrics, inspired by randomized smoothing techniques from the computer vision domain \citep{cohen2019certified}, to systematically assess the resilience of KGE models against these embedding perturbations.

Extensive experiments on widely used benchmark datasets show RKGE-D's effectiveness. The framework consistently surpasses current top KGE models, especially in tests on perturbed KGs. These results underscore the value of denoising strategies in boosting KGE robustness, pointing to a promising research direction in this field.

\section{Related Work}
\paragraph{Knowledge Graph Embedding}
KGE models use a scoring function \( f_r(h,t) \) to assess the confidence of a triple \( (h,r,t) \).
Representative geometric models like TransE \citep{transe}, RotatE \citep{rotate}, Rot-Pro \citep{rot-pro} and PairRE \citep{pairre}, HousE \citep{house} assume a relation-specific transformation brings 
$h$ close to $t$ in $n$-dimensional space. Tensor decomposition models such as DistMult \citep{distmult}, ComplEx \citep{CompLex}, TuckER \citep{TuckER}, base $f_r$ on embedding similarities of $h,r$ and $t$. Meanwhile, deep learning approaches like ConvE \citep{conve} utilize convolutional networks for feature extraction, and with the rise of graph neural networks (GNNs), GNN-based models like GAATs \citep{gaat} and NBFNet \citep{nbfnet} leverage neighboring information for knowledge representation.

\paragraph{Denoising and robustness} 
Several denoising techniques have been explored to enhance model robustness. Denoising autoencoders improves model stability by learning from corrupted data inputs \citep{vicent2010dae}. Adversarial training incorporates adversarial noise during training to make models more resilient \citep{goodfellow2015explainingharnessingadversarialexamples, madry2019deeplearningmodelsresistant}. Randomized smoothing trains models with Gaussian noise to ensure stability against input variations \citep{cohen2019certified}. These techniques collectively enhance the resilience of machine learning models in complex domains like image \citep{sahak2023denoisingdiffusionprobabilisticmodels} and large language model \citep{naacl2024robustllm}.


\section{Methodology}
\subsection{Preliminary and Notation}
Let \(\mathcal{G} = (\mathcal{E}, \mathcal{R})\) represent a knowledge graph, where \(\mathcal{E}\) is the set of entities and \(\mathcal{R}\) the set of relations. A fact in the KG is expressed as a triple \((h, r, t)\), with \(h \in \mathcal{E}\) as the head, \(r \in \mathcal{R}\) as the relation, and \(t \in \mathcal{E}\) as the tail. The energy function \(E(h, r, t)=-f_r(h,t)\) associates lower energy with higher plausibility. Energy-based KGE models aim to learn embeddings such that valid triples have lower energy than invalid ones, where \(\mathbf{h}, \mathbf{r}, \mathbf{t} \in \mathbb{R}^d\).

\subsection{Denoising as auxiliary loss}
In KGE, we create a noisy version of the dataset by randomly perturbing the triples, which involves adding Gaussian noise to the embeddings of entities. Specifically, each entity embedding \( e_i \) is perturbed as follows:

\begin{equation}
       \tilde{\mathbf{e}}_i = \mathbf{e}_i + \alpha \dot \epsilon_i, \quad \epsilon_i \sim \mathcal{N}(0, \sigma),
   \end{equation}

where \(\epsilon_i\) is Gaussian noise,  noise scale $\alpha$ is a tuneable hyperparameter, and $\sigma$ is the 99.73\%  quantile point of $|\mathbf{e}_i|$. Here we use $\mathcal{N}(0,\sigma)$ instead of $\mathcal{N}(0,1)$ because, unlike other fields like image, the input value of the image is fixed between $[0,255]$. In KGE, the range of embedding varies across different models. Moreover, the embedding of KGE models almost has outliers, making it unfeasible to use common normalization methods such as Min-Max normalization~\citep{patro2015normalization}.

The KGE model is then designed to predict these perturbations, taking the noisy triples as input and learning to output the added noise for each entity.

\paragraph{Denoising via gradient} We leverage the established link between denoising autoencoders and score matching \citep{vincent2018}, showing that the denoising objective aligns with learning the energy gradient directly from the representations of perturbed triples:
\begin{equation}
\mathbb{E}_{q_\sigma(\tilde{\mathbf{e}})} \left[ \left\| \mathcal{M}_\theta(\tilde{\mathbf{e}}) - \nabla_{\tilde{\mathbf{e}}} \log q_\sigma(\tilde{\mathbf{e}}) \right\|^2 \right],       
\end{equation}

where \(\mathcal{M}_\theta(\tilde{e})\) is the KGE model used to predict scores with noisy embeddings, and \(\nabla_{\tilde{\mathbf{e}}} \log q_\sigma(\tilde{\mathbf{e}})\) is the gradient of the noise distribution.

To be more specific, we define $\nabla_{\tilde{\mathbf{e}}} \log q_\sigma(\tilde{\mathbf{e}}) = \nabla_{\tilde{\mathbf{h}}} E(\tilde{\mathbf{h}}, \mathbf{r}, \mathbf{t})$ to be the empirical distribution \footnote{KGE typically uses the ``reverse\_relation" technique, introducing a corresponding inverse for each relation to enable bidirectional learning between entities \citep{conve}. Therefore, $\tilde{\mathbf{e}}$ is equivalent to $\tilde{\mathbf{h}}$.}. The denoising loss can be defined as follows:

\begin{equation}
    \mathcal{L}_{d} = \| \mathbf{n} - \hat{\mathbf{n}} \|^2 , \hat{\mathbf{n}} = -\nabla_{\tilde{\mathbf{h}}} E(\tilde{\mathbf{h}}, \mathbf{r}, \mathbf{t})
\end{equation}

\paragraph{Optimizing target}
Finally, the optimization goal of the model is defined as the joint loss function of the original model loss and the de-noising loss,

\begin{equation}
\mathcal{L} = \mathcal{L}_{o}+\lambda\mathcal{L}_{d},
\end{equation}

where $\mathcal{L}_{o}$ is the original loss of arbitrary backbone KGE model, and $\lambda$ is the hyperparameter used to adjust the weight between the two losses.

\subsection{Robustness Certification of KGE Models}

In the link prediction task, the input sample is represented by the query $q = (h, r, ?)$ or $(?, r, t)$, and the output entity is represented by $e$. We use $CR(\mathcal{M}, q)$ to represent the certified radius of the model $\mathcal{M}$ around $q$.

We consider the link prediction as a binary classification task, i.e., determining whether the target entity is correctly predicted.
To measure the maximum allowable perturbation to the input data while maintaining correct model output, we employ the certified radius ($CR$) as defined in \citep{cohen2019certified} and define the following definition for solving the $CR$ of the KGE models in the link prediction task. The detailed preliminary of robustness certification is introduced in Appendix \ref{app:random_smoothing}.

\begin{definition}[$CR$ in link prediction]
Suppose that a KGE model, denoted as $\mathcal{M}$, receives a query $q$ along with Gaussian noise $\epsilon\sim\mathcal{N}(0,\sigma^2)$. Furthermore, $\mathcal{M}$ has a lower bound probability of $\underline{p_T}$ with confidence $C$ for correctly outputting the entity $e_T$ when tested $n_0$ times,

if $\underline{p_T}\in(\frac{1}{2},1]$ satisfies

\begin{equation}
P(\mathcal{M}(q,\epsilon)=e_T)\geq \underline{p_T},
\end{equation}

then, CR can be expressed as
\begin{equation}
\label{eq:robust_CR}
CR(\mathcal{M}, q)=\sigma\Phi^{-1}(\underline{p_T}).
\end{equation}

For any $||\delta||_2<\sigma\Phi^{-1}(\underline{p_T})$, there is $\mathcal{M}(q,\delta)=e_T$.
\end{definition}

\subsubsection{Robustness Evaluation Metric}
\label{sec:robust_test_metric}
We adopt $ACR$ (Average Certified Radius) and $CA$ (Certified Accuracy) from ~\citep{cohen2019certified,zhai2019macer,zhang2023care}
to evaluate the robustness of the model.

$ACR$ reflects the average certified radius of the model over the test dataset. For each test triple $T_i$ and model $\mathcal{M}$, we can calculate the certified radius $CR(\mathcal{M}, T_i)$ of model $\mathcal{M}$ at triple $T_i$ according to Eq.~\ref{eq:robust_CR}. Further, $ACR$ can be expressed as:

\begin{equation}
\label{eq:robust_ACR}
ACR=\frac{1}{N}\sum_{i=1}^{N} CR(\mathcal{M}, T_i),
\end{equation}

where $N$ is the number of triples in the test dataset.

Due to the dependency of the noise standard deviation on the value of the embedding, we use the $ACR/\sigma$ to evaluate the robustness performance of the model and achieve a unified measurement standard. It can be expressed as:

\begin{equation}
\label{eq:ACR/sigma}
ACR/\sigma = \frac{1}{N} \sum_{i=1}^{N}  \frac{CR(\mathcal{M}, T_i)}{\sigma}
\end{equation}

$CA(R_p)$ reflects the proportion of the triples that has a $CR$ greater than the perturbation radius $R_p$ in the test set, and it can be expressed as:

\begin{equation}
\label{eq:robust_CA(R)}
CA(R_p) = \frac{1}{N} \sum_{i=1}^{N} \mathbf{1}[CR(\mathcal{M}, T_i)>R_p] 
\end{equation}

We define $CA(0)$ by setting \( R_p = 0 \) in Equation \eqref{eq:robust_CA(R)}, which measures the model's robustness performance. For simplicity, we denote \( CA(0) \) as \( CA \) throughout the paper.

\begin{table*}[!ht]
    \caption{Link Prediction and Robustness Validation on FB15k-237}
        \label{table:robust_link_common_metric_fb}
        \centering
        \resizebox{0.95\textwidth}{!}{
        \begin{tabular}{ll ccccc ccccc cc}
        \toprule
     & & \multicolumn{5}{c}{Link Prediction ($\alpha=2$)} &  \multicolumn{5}{c}{Link Prediction ($\alpha=5$)} & \multicolumn{2}{c}{Robustness} \\ \cmidrule(lr{.75em}){3-7} \cmidrule(lr{.75em}){8-12} \cmidrule(lr{.75em}){13-14} 
    ~ & &MRR &  MR & Hit@1 & Hit@3 & Hit@10 & MRR &  MR & Hit@1 & Hit@3 & Hit@10 & $ACR/\sigma$ & $CA$  \\ 
    \midrule
    \multirow{4}{*}{GM} &
    TransE & .262 & 266 & .176  & .291 & .431 & .133 & 649 & .077 & .144 & .244 & .321 & .194 \\ 
     & RotatE  & .286 & 244 & .201 & .314 & .456 & .161 & 448 & .096 &  .173 & .288 & .333 & .203 \\
    & PairRE & .275 & 289 & .194  & .300 & .438 & .194 & 730 & .131 & .208 & .321 & .333 & .203 \\
    & HousE & .271 & 265 & .187 & .300 & .441 & .210 & 397 & .136 & .231 & .361 & .572 & .263 \\
     
    \midrule
     \multirow{3}{*}{TD} & DistMult & .282 & 220 & .197 & .307 & .455 & .189 & 579 & .120 & .205 & .326 & .396 & .208 \\
    & ComplEx   & .281 & 273 & .193 & .306 & .457 & .161 & 883 & .099 & .173 & .285 &  .367 &	.210\\
    & TuckER   & .255 & 474 & .176  & .276 & .410 & .141 & 2141 & .092 & .150 & .237 & .520 & .253\\
   
    \midrule
     \multirow{2}{*}{DL} & 
    ConvE & .174 & 323 & .121 & .186 & .276 & .050 & 574 & .031 & .047 & .079 & .229 & .156 \\
    & HConvRot & .197 & 254 & .134 & .217 & .322 & .078 & 495 & .044 & .082 & .141 &   .219 & .166 \\
    \midrule
   \multirow{2}{*}{Ours}  & TuckER-D & \underline{.294} & \textbf{197} & \underline{.214} & \underline{.319}  & \underline{.451} & \textbf{.286} & \textbf{200} & \textbf{.198} & \textbf{.301} & \underline{.401}  & \underline{.526} & \underline{.253} \\

   & HousE-D & \textbf{.302} & \underline{217} & \textbf{.214} & \textbf{.334} & \textbf{.476} & \underline{.263} & \underline{275} & \underline{.179} & \underline{.292} & \textbf{.430} & \textbf{.578} & \textbf{.266}  \\

    \bottomrule
        \end{tabular}
        }
\end{table*}

\section{Experiments and Analysis}

\subsection{Experimental Setting}

\subsubsection{Datasets}
We evaluate the performance of our proposed RKGE-D framework in the link prediction task using a well-known benchmark dataset
FB15k-237, which is derived from Freebase, with 237 relations and fewer inverse relations \citep{conve}.

\subsubsection{Hyperparameters}

For the introduced hyperparameters, we use grid search of hyperparameters to perform model enhancement under the RKGE-D framework: the scale of the training noise $\alpha\in\{0.1,0.2,0.5,1.0\}$, weight of $\mathcal{L}_d$, $\lambda\in\{0.1,0.2,0.5,1.0\}$. Moreover, all the robust metrics are certified with testing times $n_0=1,000$ and confidence $C=99.9\%$.

\subsubsection{Baselines}
For the baseline KGE models, we select geometric models (GM) including TransE~\citep{transe}, RotatE~\citep{rotate}, PairRE~\citep{pairre}, and Rot-Pro~\citep{rot-pro}, tensor decomposition (TD) models such as DistMult~\citep{distmult} and ComplEx~\citep{CompLex}, and deep learning (DL) models ConvE~\citep{conve}, HConvRot \citep{convrot} Robust training is performed using the RKGE-D framework, applied to two state-of-the-art models: HousE and TuckER.


\subsubsection{Evaluation}
In link prediction tasks, we use the evaluation method from ConvE, generating two queries for each test triple to predict both head and tail entities. We rank all entities as potential targets based on model scores, following the filtering settings from TransE, where known triples in the dataset are omitted from rankings. We assess performance using metrics such as MRR, MR, and Hit@$k$ ($k=1,3,10$),  Higher MRR, Hit@$k$ and lower MR indicate better results.

When evaluating the robust metric, we use the robust metric $ACR/\sigma$ and $CA$ proposed in section~\ref{sec:robust_test_metric} to measure the robustness performance of the models.

\subsection{Main Results}
Table~\ref{table:robust_link_common_metric_fb} shows the link prediction result on perturbed entity embedding with noise scale $\alpha=2,5$, and the robustness validation metric of 3 types of KGE methods. Better results are in \textbf{bold}. Note that, due to HousE’s strong stability against perturbations, the effect of RKGE-D is not evident under small noise levels. Therefore, we applied $\alpha=100,150$ specifically to HousE to verify the effectiveness of the denoising mechanism. 

We observe notable robustness differences among popular KGE methods. Specifically, deep learning-based models are more vulnerable to perturbations, and models with superior generalization capabilities tend to exhibit greater robustness.

We can also see that models using the RKGE-D framework significantly outperform their backbone counterparts. This demonstrates that our noise-based robust training method effectively enhances model robustness.

\subsection{Hyperparameter sensitivity}
\begin{figure}[htb]
\centering
\subfloat[$\alpha$]{
   
    \label{fig:conve_alpha}
    \centering
    \includegraphics[width=0.25\textwidth]{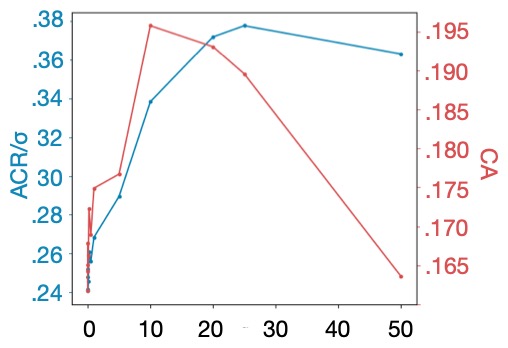}

}
\subfloat[$\lambda$]{
    \label{fig:conve_lambda}
    \centering
    \includegraphics[width=0.25\textwidth]{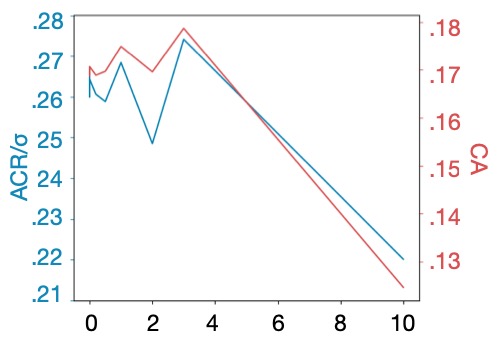}
}
\caption{Hyperparameter sensitivity}
\label{fig:hyperparameter}
\end{figure}

Figure \ref{fig:hyperparameter} shows that during the training process, how the noise scale $\alpha$ and weight of denoising loss $\lambda$ affect the model performance. 

We can see both excessively small or large values for $\alpha$ and $\lambda$ during training lead to performance degradation. Notably, $\alpha=0$ or $\lambda=0$ reduces the robustness, affirming the effectiveness of the proposed RKGE-D.

\section{Conclusion}
In this paper, we introduced a robust denoising framework for KGE training and robustness validation against embedding-level perturbations. Extensive experiments demonstrate the superiority of RKGE-D over state-of-the-art models in noisy embedding scenarios and provide a comparative analysis of robustness with existing KGE methods.

\section*{Limitation}
We also conduct link prediction tests on unperturbed data.
While our proposed framework also shows some improvement (about +0.4\%) without noise addition, the enhancement is not substantial. This limited impact might be attributed to the inherent challenges associated with randomized smoothing, as highlighted by \citep{naacl2024robustllm}, directly applying randomized smoothing to models results in unsatisfactory robustness. This is largely due to the fact that randomized smoothing adds noise to the input, and the enhanced robustness of the final model critically hinges on how well it can handle these noise-corrupted data. In future work, we aim to develop methods to optimize model handling of clean data while maintaining robustness, potentially through adaptive noise management or advanced noise simulation techniques.

\section*{Ethical Considerations}
In developing a robust denoising framework for KGE, we address a critical limitation in the resilience of KGE models to noisy or adversarial data, which may benefit the reliability of AI systems in sensitive domains such as healthcare, finance, and legal reasoning in the future. However, a key ethical consideration is that the reliance on existing KGE models may unintentionally perpetuate biases, which could skew results in undesirable ways. To mitigate this risk, it is crucial to carefully monitor data sources and ensure transparency in how the model processes and adjusts for potential biases. Future iterations of this framework should prioritize fairness by incorporating debiasing techniques and informing users about any limitations.

\bibliography{custom}

\newpage

\appendix

\section{Randomized Smoothing}
\label{app:random_smoothing}

Randomized smoothing \citep{cohen2019certified} posits that the certified robust radius can be solved based on a smoothed classifier \( g \) constructed from a base classifier \( f \). The classifier \( g \) is constructed as follows: for an input sample \( x \), Gaussian noise \( \epsilon \in \mathcal{N}(0, \sigma) \) is added as perturbation \( \delta \), and \( g \) outputs the class most likely predicted by \( f \):

$$
g(x) = \mathop{\arg\max}\limits_{c\in\mathcal{C}} P(f(x,\epsilon) = c),
$$
where \( c \) is a class and \( \mathcal{C} \) represents the class set. \( f(x, \epsilon) \) denotes the classifier \( f \)'s output when input \( x \) is perturbed by noise \( \epsilon \).

Then, the certified robust radius is solved using the smoothed classifier \( g \). For binary classification, if \( g \) outputs the correct class \( c_T \) with probability \( p_T \) and the wrong class \( c_F \) with probability \( p_F = 1 - p_T \), where \( p_T > p_F \), the robust radius of \( g \) at \( x \) with respect to \( \ell_2 \)-norm is given by:

$$
CR = \sigma\Phi^{-1}(p_T),
$$
where \( \Phi^{-1} \) is the inverse of the cumulative distribution function of the standard Gaussian distribution. Since the probability \( p_T \) varies with different noise \( \epsilon \), the certified robust radius derived from this may not always be valid.

To address this, a lower bound \( \underline{p_T} \) is introduced to achieve a high-confidence certified robust radius. If the lower bound \( \underline{p_T} \in (0.5, 1] \) satisfies

$$
P(f(x, \epsilon) = c_T) = p_T \ge \underline{p_T},
$$

then substituting \( \underline{p_T} \) for \( p_T \), the above conclusion still holds, i.e., \( g \) remains robust within the \( \ell_2 \)-radius \( CR = \sigma\Phi^{-1}(\underline{p_T}) \).

The lower bound \( \underline{p_T} \) is obtained as follows: perform \( n_0 \) perturbations with noise \( \epsilon \), count the number of times the classifier \( f \) outputs the correct class \( c_T \), and use the Lower Confidence Bound (LCB) function to compute the lower bound \( \underline{p_T} = \text{LCB}(n_0, \text{count}, C) \), where the LCB function returns the one-sided confidence interval of the binomial parameter \( \underline{p_T} \) with confidence level \( C \).

At high confidence \( C \), the gap between the certified robustness of the base classifier \( f \) and its smoothed version \( g \) becomes negligible, allowing the certified robust radius of classifier \( f \) to be derived from that of the smoothed classifier \( g \). Thus:

$$
CR_f = CR_g = \sigma\Phi^{-1}(\underline{p_T}).
$$

\begin{table*}[ht]
    \caption{RKGE-D performance on FB15k-237 for multi-hop reasoning tasks.}
    \centering    \label{table:robust_reason_common_metric}
    \resizebox{0.9\textwidth}{!}{
    \begin{tabular}{cc ccc ccc ccc ccc}
    \toprule
        \multirow{2}{*}{} & \multirow{2}{*}{}  & \multicolumn{3}{c}{MRR} & \multicolumn{3}{c}{Hit@1} & \multicolumn{3}{c}{HIT@3} & \multicolumn{3}{c}{HIT@10} \\ 
        \cmidrule(lr{.75em}){3-5} \cmidrule(lr{.75em}){6-8} \cmidrule(lr{.75em}){9-11} \cmidrule(lr{.75em}){12-14}
        ~ & ~  & 1p & 2p & 3p & 1p & 2p & 3p & 1p & 2p & 3p & 1p & 2p & 3p  \\ \midrule
        \multirow{8}{*}{GM} & TransE &  34.2 & 6.7 & 5.4 & 24.4 & 2.9 & 2.4 & 38.3 & 6.5 & 5.1 & 53.5 & 13.7 & 10.8  \\
        ~ & TransE-D & \textbf{34.6} & \textbf{6.8} & 5.3 & \textbf{25.0} & \textbf{3.2} & \textbf{2.4} & \textbf{38.5} & 6.5 & \textbf{5.1} & 53.4 & \textbf{13.7} & 10.4 \\ \cline{2-14}
        ~ & RotatE & 43.8 & 8.9 & 5.6 & 33.1 & 4.5 & 2.4 & 48.7 & 8.8 & 5.3 & 65.5 & 17.0 & 11.2  \\
        ~ & RotatE-D & {43.7} & \textbf{9.1} & \textbf{5.7} & {32.8} & \textbf{4.5} & \textbf{2.4} & \textbf{48.8} & \textbf{9.0} & \textbf{5.3} & {65.4} & \textbf{17.3} & \textbf{11.5}  \\ \cline{2-14}
        ~ & PairRE  & 44.4 & 9.7 & 7.1 & 33.9 & 5.2 & 3.5 & 49.4 & 9.7 & 6.9 & 65.1 & 18.0 & 13.6  \\
        ~ & PairRE-D & \textbf{45.0} & \textbf{10.0} & \textbf{7.1} & \textbf{34.6} & \textbf{5.3} & \textbf{3.5} & \textbf{49.9} & \textbf{9.9} & \textbf{6.9} & \textbf{66.1} & \textbf{19.0} & \textbf{14.3}  \\ \cline{2-14}
        ~ & Rot-Pro & 42.6 & 7.8 & 5.0 & 32.6 & 3.9 & 2.2 & 47.3 & 8.0 & 5.1 & 62.7 & 15.1 & 10.2 \\
        ~ & Rot-Pro-D & \textbf{43.9} & \textbf{9.0} & \textbf{5.8} & \textbf{33.1} & \textbf{4.5} & \textbf{2.5} & \textbf{48.9} & \textbf{8.8} & \textbf{5.5} & \textbf{65.7} & \textbf{17.8} & \textbf{11.7} \\ \midrule

        \multirow{4}{*}{TD} & ComplEx & 20.1 & 4.4 & 2.1 & 11.4 & 1.9 & 0.9 & 21.5 & 4.2 & 1.9 & 38.7 & 9.0 & 4.1 \\
        ~ & ComplEx-D & \textbf{20.4} & 44.2 & \textbf{2.2} & \textbf{12.1} & \textbf{2.2} & \textbf{1.0} & \textbf{22.1} & 4.1 & \textbf{2.0} & 37.9 & 8.0 & 4.1  \\ \cline{2-14}
        ~ & DistMult & 27.1 & 6.3 & 3.4 & 16.6 & 3.0 & 1.5 & 30.4 & 5.9 & 3.1 & 49.6 & 12.6 & 6.6 \\
        ~ & DistMult-D  & \textbf{27.9} & \textbf{6.5} & \textbf{3.5} & \textbf{17.2} & \textbf{3.1} & \textbf{1.6} & \textbf{31.2} & \textbf{6.2} & \textbf{3.2} & \textbf{51.2} & \textbf{13.0} & \textbf{6.7}  \\ \midrule

        \multirow{6}{*}{DL} & ConvE & 40.4 & 7.2 & 5.1 & 30.3 & 3.7 & 2.4 & 44.4 & 7.1 & 4.9 & 61.1 & 13.9 & 9.9  \\
        ~ & ConvE-D  & \textbf{41.9} & \textbf{7.7} & \textbf{5.6} & \textbf{31.8} & \textbf{3.9} & \textbf{2.8} & \textbf{46.3} & \textbf{7.5} & \textbf{5.2} & \textbf{62.2} & \textbf{15.0} & \textbf{10.8}  \\ \cline{2-14}
        ~ & HConvRot  & 41.7 & 5.0 & 2.3 & 31.6 & 2.4 & 1.1 & 46.6 & 4.9 & 2.1 & 61.6 & 9.6 & 4.2  \\
        ~ & HConvRot-D & \textbf{42.1} & \textbf{5.3} & \textbf{2.4} & \textbf{32.1} & \textbf{2.7} & \textbf{1.2} & \textbf{46.6} & \textbf{5.1} & \textbf{2.3} & \textbf{62.1} & \textbf{10.1} & \textbf{4.5} \\ \cline{2-14}
         & KBGAT  & 34.2 & 7.0 & 5.7 & 24.6 & 3.1 & 2.4 & 38.0 & 7.0 & 5.5 & 53.1 & 13.9 & 11.2 \\
        ~ & KBGAT-D & \textbf{36.1} & \textbf{7.6} & \textbf{5.7} & \textbf{25.9} & \textbf{3.6} & \textbf{2.5} & \textbf{40.5} & \textbf{7.5} & \textbf{5.6} & \textbf{56.4} & \textbf{15.0} & \textbf{11.4} \\
    \bottomrule
    \end{tabular}
    }          
\end{table*}

\section{Experiment}
\subsection{Computational Experiments}
All our experiments were conducted on a server with Intel Xeon Gold 2.40@GHz CPU and NVIDIA A100 40GB GPU.
Each model is trained using one GPU, which takes 6 GPU hours on average.


\subsection{Downstream Task on Multi-hop Reasoning}
As mentioned in Figure \ref{fig:robust_kge}, the error caused by inefficient robustness will propagate through link prediction inference as the number of inference hops increases, severely affecting the results of downstream tasks such as multi-hop reasoning. Therefore, we further validate the performance of the proposed RKGE-D framework in multi-hop reasoning. 

Multi-hop reasoning refers to inferring indirect relationships between two entities by traversing multiple relational paths within the knowledge graph. Unlike simple one-hop reasoning, multi-hop reasoning requires the model to understand complex path structures and relationships between intermediate nodes. This task aims to deduce implicit information in the graph by reasoning across multiple relational chains, which plays a crucial role in answering complex questions, discovering hidden knowledge, and enhancing graph completion capabilities. However, it also places higher demands on the model's expressiveness and robustness.

\paragraph{Evaluation}
We follow the evaluation method in BetaE~\citep{ren2020beta} to evaluate the results of the model on various query types, across 1p, 2p, 3p (projection), and 2i, 3i, ip, pi (intersection and union queries).

\paragraph{Multi-hop reasoning result}
Table \ref{table:robust_reason_common_metric} shows the general metric results of multi-hop reasoning tasks for nine benchmark models on the FB15k-237 dataset, both before and after applying the RKGE-D framework proposed in this chapter. The multi-hop reasoning abilities of most geometric and tensor decomposition models show only minor improvements, whereas CNN and GNN models demonstrate more significant enhancements in their multi-hop reasoning performance.

\paragraph{Case Study}

\begin{figure}[h]
    \centering
    \includegraphics[width=\linewidth]{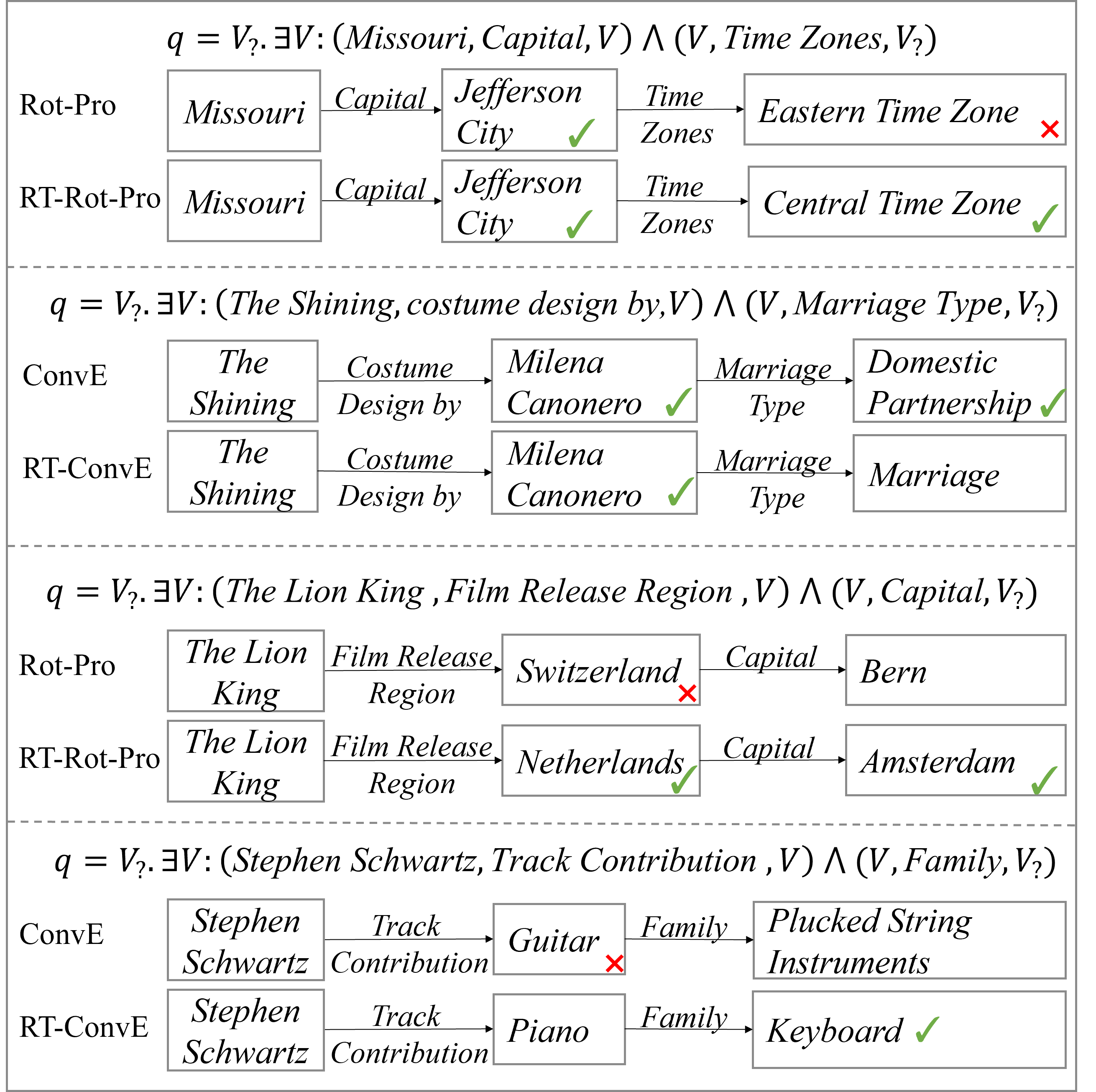}
    \caption{Case Study}
    \label{fig:robust_case_study}
\end{figure}

In this section, we evaluate the performance of the RKGE-D framework in downstream multi-hop reasoning tasks on KGs. Multi-hop reasoning involves deducing indirect relations between entities by traversing multiple relational paths. Unlike single-hop reasoning, it requires models to understand complex path structures and intermediate relations, making it crucial for answering complex questions and enhancing knowledge graph completion. This task is challenging, demanding robust models that can infer hidden knowledge from multi-step relational chains.

we select Rot-Pro and ConvE to generate several cases on FB15k-237 and
show the cases in Figure~\ref{fig:robust_case_study}, aiming to conduct an in-depth analysis of the robustness framework RKGE-D. 

Taking the first case as an example, the meaning of the query is "What is the time zone of Missouri's capital?"
In the second hop inference, Rot-pro-D ranks the correct answer Eastern Time Zone first by predicted score ranking, while Rot-Pro ranks Central Time Zone first. 
It provides a more intuitive demonstration of how the RKGE-D framework enhances the multi-hop reasoning capability of KGE models.

\end{document}